\documentclass{article}
\pdfoutput=1

\usepackage[final]{tech}

\usepackage[T1]{fontenc}    % use 8-bit T1 fonts
\usepackage{booktabs}       % professional-quality tables
\usepackage{nicefrac}       % compact symbols for 1/2, etc.
\usepackage{microtype}      % microtypography
\usepackage{hyperref}
\usepackage{url}
\usepackage{graphicx}
\usepackage{subfigure}
\usepackage{float}
\usepackage{enumerate} 
\usepackage{bbm}
\usepackage{amsmath,amsfonts,amssymb,amsthm,commath,dsfont}
\usepackage{algorithm,algorithmicx,algpseudocode}
\usepackage{mathtools}
\usepackage{mathabx}
\usepackage{nicefrac}

\def\ddefloop#1{\ifx\ddefloop#1\else\ddef{#1}\expandafter\ddefloop\fi}
\def\ddef#1{\expandafter\def\csname bb#1\endcsname{\ensuremath{\mathbb{#1}}}}
\ddefloop ABCDEFGHIJKLMNOPQRSTUVWXYZ\ddefloop
\def\ddef#1{\expandafter\def\csname c#1\endcsname{\ensuremath{\mathcal{#1}}}}
\ddefloop ABCDEFGHIJKLMNOPQRSTUVWXYZ\ddefloop

\def\R{\mathbb{R}}

\def\1{\mathds{1}}

\def\rmse{\textup{RMSE}}
\def\se{\textup{SE}}

\newcommand{\ip}[2]{\left\langle #1, #2 \right \rangle}

\theoremstyle{definition}
\newtheorem{definition}{Definition}
\theoremstyle{remark}

\theoremstyle{assumption}
\newtheorem{assumption}{Assumption}

\DeclarePairedDelimiter\floor{\lfloor}{\rfloor}

\def\Lsize{\hbox{\space \raise-2mm\hbox{$\textstyle \L \atop \scriptstyle {m\times 3n}$} \space}}
\def\Ssize{\hbox{\space \raise-2mm\hbox{$\textstyle \S \atop \scriptstyle {m\times 3n}$} \space}}
\def\Osize{\hbox{\space \raise-2mm\hbox{$\textstyle \Ome \atop \scriptstyle {m\times 3n}$} \space}}
\def\Tsize{\hbox{\space \raise-2mm\hbox{$\textstyle \T \atop \scriptstyle {3n\times n}$} \space}}
\def\Bsize{\hbox{\space \raise-2mm\hbox{$\textstyle \B \atop \scriptstyle {m\times n}$} \space}}

\title{Attack RMSE Leaderboard: An Introduction and Case Study}

\author{
  Cong Xie \\
  Department of Computer Science\\
  University of Illinois at Urbana-Champaign \\
  \texttt{cx2@illinois.edu} \\
}

\begin{document}

\maketitle

\begin{abstract}
In this manuscript, we briefly introduce several tricks to climb the leaderboards which use RMSE for evaluation without exploiting any training data. 
\end{abstract}

\section{Introduction}
\underline{R}oot-\underline{M}ean-\underline{S}quared-\underline{E}rror~(RMSE) is no longer widely used for evaluation in data challenges because of its easy-to-hack properties. To be more specific, given any vector $\hat{y} \in \R^n$ and the true labels $y \in \R^n$, where $n$ is the number of desired label, we can easily obtain the inner product $\ip{\hat{y}}{y}$, which can be used to extract the information of the true labels $y$.

There are other kinds of vulnerable evaluations such as log-loss~\citep{Whitehill2017ClimbingTK}. For classification problems, there is also Monte-Carlo style attacks such as \cite{MrtzCompetition}.

\section{Notations}
Denote the desired vector of the true labels as $y \in \R^n$, where $n$ is the number of labels. $\hat{y} \in \R^n$ is the vector of submitted labels. After each submission, the data challenge platform~(such as Kaggle) will evaluate the submitted labels and provide the RMSE oracle which is defined as follows:
\begin{definition}
\underline{R}oot-\underline{M}ean-\underline{S}quared-\underline{E}rror~(RMSE): 
\begin{align*}
\rmse(\hat{y}) = \sqrt{\frac{1}{n} \sum_{i=1}^n (\hat{y}_i - y_i)^2} = \sqrt{\frac{1}{n} \|\hat{y} - y\|^2},
\end{align*}
where $\|\cdot\|$ is the $\ell_2$-norm.
\end{definition}
For simplicity, we denote $y^2 = \|y\|^2$, $\hat{y}^2 = \|\hat{y}\|^2$, $\se(\hat{y}) =  \|\hat{y} - y\|^2$. Furthermore, we denote the mean values as $m = \frac{1}{n} \sum_{i=1}^n y_i$, $\hat{m} = \frac{1}{n} \sum_{i=1}^n \hat{y}$, and the segmental mean values as $m_{i:j} = \frac{1}{j-i+1} \sum_{k=i}^j y_k$, $\hat{m}_{i:j} = \frac{1}{j-i+1} \sum_{k=i}^j \hat{y}$.

\section{Assumptions}
Globally, we take the following assumptions:
\begin{assumption}
The labels are upper-bounded and lower-bounded: $y_i \in [a, b]$ for $\forall i \in [n]$.
\end{assumption}
The above assumption is easily satisfied in most data challenges. The constants $a$ and $b$ can be easily inferred/estimated from the background information of the challenges.

\section{Methodology}
In this section, we introduce several tricks to climb the leaderboards. 

For most of the tricks, the basic ideas are very similar: we first submit $\hat{y} = 0$, the RMSE oracle $\sqrt{\frac{1}{n} \|\hat{y} - y\|^2} = \sqrt{\frac{1}{n} \|0 - y\|^2} = \sqrt{\frac{1}{n} \|y\|^2}$ will then give us $y^2$. Once we obtain $y^2$, for any $\hat{y} \in \R^n$, we can obtain the inner product $\ip{\hat{y}}{y}$ via the decomposition $\se(\hat{y}) = \hat{y}^2 + y^2 - 2\ip{\hat{y}}{y}$. Note that $y^2$ and $\hat{y}^2$ are known, and $\se(\hat{y})$ can be inferred from the RMSE oracle. Thus, we obtain $\ip{\hat{y}}{y} = \frac{1}{2} [\hat{y}^2 + y^2 - \se(\hat{y})]$. In the rest of this manuscript, we assume that $\ip{\hat{y}}{y}$ is already known for any $\hat{y}$.

\subsection{Mean-value Attack}
\label{sec:mean_attack}
The idea is very simple. For any segment of indices $i, \ldots, j$, we take $\hat{y}' = [0, \ldots, \underbrace{1, \ldots, 1}_{i, \ldots, j}, \ldots, 0]$. Hence, we obtain the mean value $m_{i:j} = \frac{1}{j-i+1} \sum_{k=i}^j y_k = \frac{1}{j-i+1} \ip{\hat{y}'}{y}$. Then, given any submission $\hat{y}$, we improve the evaluation by using Algorithm~\ref{alg:mean_adj}.

\begin{algorithm}[H]
\caption{Mean-value improvement}
\begin{algorithmic}
\Require Any segment of indices $\{i, \ldots, j\}$, the corresponding submitted labels $\tilde{y} = \hat{y}_{i:j}$ and the mean value $\tilde{m} = m_{i:j}$, and the lower bound $a$ and upper bound $b$. 
\Ensure The improved segment of submission $\bar{y}$.
\State Use any solver~(e.g. \cite{quadprog}) to solve the following quadratic programming with constraints:
\begin{align*}
&\min_{\bar{y}} \frac{1}{2} \|\bar{y} - \tilde{y}\|^2, \\
s.t.\quad & \1^\top \bar{y} = (j-i+1)\tilde{m}, \\
& a \1 \leq \bar{y} \leq b \1.
\end{align*}
\end{algorithmic}
\label{alg:mean_adj}
\end{algorithm}

Note that when the size of the segment is $1$, the inner product is the exact value of the corresponding true label. Furthermore, we obtain better evaluation with more submissions for different segments. 

\subsection{Linear-regression Attack}
Assume that the budget of submissions~(the remaining number of submissions) is $r \leq n$. We somehow generate a matrix $A \in \R^{n \times r}$ composed of $r$ columns of bases, where each column is linearly independent or orthogonal to one another. Note that some or all of the columns of $A$ can be different submissions. We can generate a submission better than any of these submissions, which is a linear combination of these columns $Ax$, where $x \in \R^r$. The optimal $x$ is obtained by solving the following least-square linear regression problem:
\begin{align*}
\min_x \frac{1}{2} \|Ax - y\|^2.
\end{align*}
Because we have the full control of the matrix $A$, we can easily make it of full rank~(e.g. with linearly independent columns), which implies that $A^\top A$ is invertible. Note that the above regression problem has the closed-form solution: $x = \left( A^\top A \right)^{-1} A^\top y$. The $j$th element of the vector $A^\top y$ is simply the inner product $\ip{A_{:, j}}{y}$, which can be easily obtained as mentioned before, where $A_{:, j}$ is the $j$th column of $A$. Hence, we generate the new submission $Ax$.

Furthermore, we can easily infer the RMSE of the new submission, which is $\sqrt{\frac{1}{n} \left[ x^\top A^\top Ax + y^2 - 2 x^\top (A^\top y) \right] }$, where $x$, $A$, $y$, and $A^\top y$ are all known. 

\subsection{Finite-label Attack}
Ideally, this kind of attack can generate perfect labels~(exactly $y$) with one single submission~(additional to the submission of obtaining $y^2$). However, such attack also requires much stronger conditions as follows:
\begin{assumption}
There are only finite number of different values for the true labels, which means that the labels are discrete.
\end{assumption}
Together with the assumption that the labels are bounded, we can easily transform the original problem into a new problem where the labels are integers lying in the range $[0, c)$, where $c$ is also an integer. Then, we take $\hat{y} = [1, c, c^2, \ldots, c^{n-1}]$. In other words, $\hat{y}_i = c^{i-1}$ for $\forall i \in [n]$. Using the resulting inner product $\ip{\hat{y}}{y}$, we can obtain all the true labels via Algorithm~\ref{alg:finite}.

\begin{algorithm}[H]
\caption{Perfect submission}
\begin{algorithmic}
\Require $\hat{y} = [1, c, c^2, \ldots, c^{n-1}]$, $p = \ip{\hat{y}}{y}$.
\Ensure Transformed true labels $y'$.
\For{$i = 1, \ldots, n$}
\State $y'_i \leftarrow \mod(p, c)$.
\State $p \leftarrow \floor*{p / c}$.
\EndFor
\end{algorithmic}
\label{alg:finite}
\end{algorithm}

Note that this attack also requires that the $c$ is small enough so that $c^n$ does not overflow. However, we can also infer a segment of length $n' < n$ of the labels so that $c^{n'}$ does not overflow. 

\subparagraph{Case study: Restaurant Revenue Prediction~\citep{RestaurantCompetition}}
In this data challenge, any true label satisfies $y_i \in \{0, 1\}$. And, the submission can be any real number. Thus, we simply take $c = 2$. Then, the perfect submission can be obtained as follows:
\begin{enumerate}
\item Submit $\hat{y} = 0$. Obtain $y^2$.
\item Submit $\hat{y} = [1, 2, 2^2, \ldots, 2^{n-1}]$. Obtain $p = \ip{\hat{y}}{y}$.
\item Use Algorithm~\ref{alg:finite} to obtain the perfect submission.
\end{enumerate}
Thus, we only need 2 submissions to get ranked as top-1~(0-RMSE) of the leaderboard!

\section{Conclusion}
RMSE as evaluation is highly vulnerable. Enjoy climbing the leaderboard!

\bibliography{rmse}
\bibliographystyle{plainnat}

\end{document}